\documentclass[letterpaper, 10 pt, conference]{ieeeconf}  

\IEEEoverridecommandlockouts                              

\usepackage{hyperref}
\usepackage{graphics} 
\usepackage{epsfig} 
\usepackage{mathptmx} 
\usepackage{times} 
\usepackage{amsmath} 
\usepackage{amssymb}  
\usepackage{xcolor}
\usepackage{multirow}
\usepackage{booktabs}
\usepackage{siunitx}
\usepackage[format=plain]{caption}
\usepackage{cite}

\usepackage{subcaption}
\usepackage{booktabs}
\usepackage{threeparttable}
\usepackage{bm}
\usepackage{multirow}

\usepackage{makecell}
\usepackage{threeparttable}
\usepackage{amssymb}

\let\labelindent\relax
\usepackage{enumitem}

\usepackage{hyperref}
\usepackage{xcolor}
\usepackage{comment}

\title{\LARGE \bf
Phase-Specific Augmented Reality Guidance for Microscopic Cataract Surgery Using Long-Short Spatiotemporal Aggregation Transformer
}

\author{Puxun Tu$^{1}$, Hongfei Ye$^{2}$, Jeff Young$^{3}$, Peiquan Zhao$^{2}$, Ce Zheng$^{2}$, 
Xiaoyi Jiang$^{4}$, Xiaojun Chen$^{1}$ 
\thanks{$^{1}$the Institute of Biomedical Manufacturing and Life Quality Engineering, School of Mechanical Engineering, Shanghai Jiao Tong University, China,
        {\tt\small xiaojunchen@sjtu.edu.cn.} }%
\thanks{$^{2}$the Department of Ophthalmology, Xinhua Hospital Affiliated to Shanghai Jiao Tong University School of Medicine, Shanghai, China,
        {\tt\small zhengce@xinhuamed.com.cn.} }%
\thanks{$^{3}$the Department of Bioengineering, Rice University, Houston, USA.}%
\thanks{$^{4}$the Faculty of Mathematics and Computer Science, University of Münster, Münster, Germany.}%
}

\begin{document}

\maketitle
\thispagestyle{empty}
\pagestyle{empty}

\begin{abstract}
Phacoemulsification cataract surgery (PCS) is a routine procedure conducted using a surgical microscope, heavily reliant on the skill of the ophthalmologist. While existing PCS guidance systems extract valuable information from surgical microscopic videos to enhance intraoperative proficiency, they suffer from non-phase-specific guidance, leading to redundant visual information. In this study, our major contribution is the development of a novel phase-specific augmented reality (AR) guidance system, which offers tailored AR information corresponding to the recognized surgical phase. Leveraging the inherent quasi-standardized nature of PCS procedures, we propose a two-stage surgical microscopic video recognition network. In the first stage, we implement a multi-task learning structure to segment the surgical limbus region and extract limbus region-focused spatial feature for each frame. In the second stage, we propose the long-short spatiotemporal aggregation transformer (LS-SAT) network to model local fine-grained and global temporal relationships, and combine the extracted spatial features to recognize the current surgical phase. Additionally, we collaborate closely with ophthalmologists to design AR visual cues by utilizing techniques such as limbus ellipse fitting and regional restricted normal cross-correlation rotation computation. We evaluated the network on publicly available and in-house datasets, with comparison results demonstrating its superior performance compared to related works. Ablation results further validated the effectiveness of the limbus region-focused spatial feature extractor and the combination of temporal features. Furthermore, the developed system was evaluated in a clinical setup, with results indicating acceptable accuracy and real-time performance, underscoring its potential for clinical applications. 

\end{abstract}

\section{Introduction}
\label{introduction}
Cataract remains the leading cause of blindness worldwide, and phacoemulsification cataract surgery (PCS) has emerged as the established standard of care for its treatment. PCS follows a quasi-standardized procedure involving specific surgical phases \cite{RN661}, enabling the removal of the cataract and the placement of an intraocular lens (IOL) to restore visual acuity. The procedure is typically performed using a surgical microscope, which offers an enhanced view of the surgical field with magnification, brightness, and clarity. However, the success of PCS is highly reliant on the surgical skills of the ophthalmologists, and statistical evidence highlights significant differences in complication rates among ophthalmologists with varying levels of seniority and experience \cite{RN646}.

In most cases, the surgical microscope used in PCS is equipped with a camera that transmits the surgical field to an external screen, allowing for intraoperative monitoring and procedure recording. The microscopic video contains rich spatiotemporal information, presenting an exceptional opportunity to develop surgical video recognition methods. These methods can extract valuable intraoperative information, such as delineating key anatomical boundaries \cite{RN645}, detecting surgical instruments \cite{RN658}, and computing the rotation \cite{RN648}. These extracted details can then be overlaid on a 2D/3D screen or the microscopic eyepiece, creating an augmented reality (AR) scene to enhance ophthalmologist's intraoperative skills \cite{RN653}.

\begin{figure}[!t]
	\centerline{\includegraphics[width=\columnwidth]{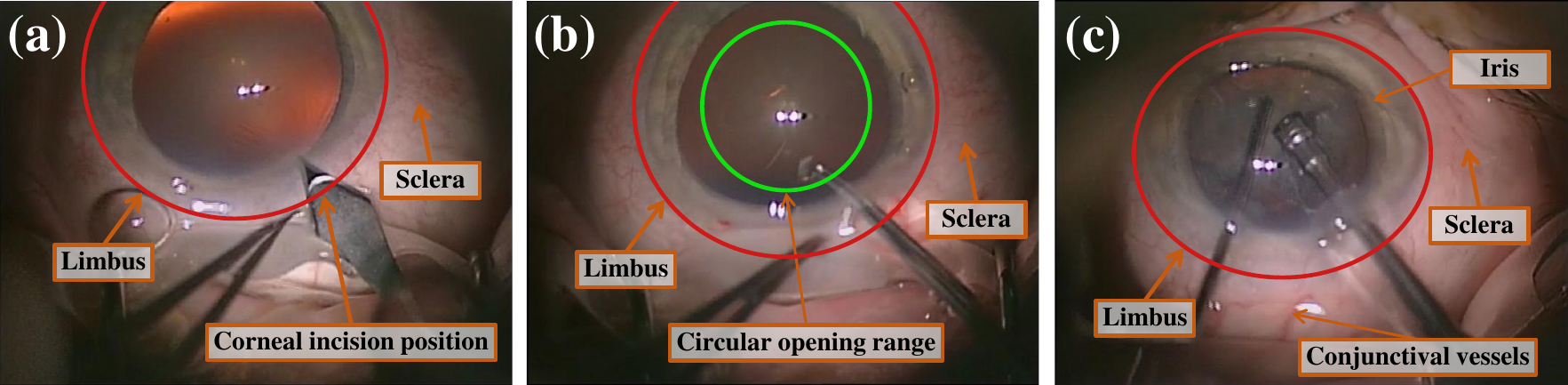}}
	\caption{Three representative surgical phases in PCS: (a) the incision phase, (b) the capsulorhexis phase, and (c) the phacoemulsification phase. Ophthalmologists adjust their focus position according to the different surgical phases. Additionally, the surgical limbus region displays distinct appearances during various surgical phases.}
	\label{fig1}
\end{figure}

Several intraoperative guidance systems for ophthalmic surgery utilizing microscopic video recognition have been proposed \cite{RN648}, \cite{RN657}, \cite{RN647}. Despite their clinical significance, certain limitations hinder their implementation in PCS. Firstly, these systems cannot provide phase-specific intraoperative AR information for ophthalmologists, leading to the issue of visual redundancy. In clinical practice, the significance of phase-specific guidance becomes evident, as ophthalmologists hold varying expectations for augmented visual information at different surgical phases. For example, during the incision phase, their focus is on the position of the corneal incision site (Fig. \ref{fig1} (a)), whereas during the capsulorhexis phase, they prioritize assessing the circular opening range of the capsule (Fig. \ref{fig1} (b)). If all AR information is provided uniformly across all surgical phases, the presence of redundant visual information could divert their attention and potentially lead to surgical complications. Secondly, existing systems process surgical videos in a frame-wise manner, enabling real-time processing but resulting in the loss of crucial temporal information. These observations motivate us to develop our phase-specific intraoperative guidance system that offers ophthalmologists distinct AR information tailored to different surgical phases.

The key technology in developing the phase-specific AR guidance system is real-time recognition of the surgical phase from microscopic video. 
While various methods have been proposed for surgical phase recognition \cite{RN660} \cite{RN654} \cite{RN651} \cite{jin2022trans} \cite{RN642}, none of them utilizes the recognized surgical phase for intraoperative AR guidance. These methods typically adopt a two-stage framework, with the first stage involving the extraction of spatial features and the second stage employing these features for temporal feature aggregation. The spatial feature extractor is typically trained in a frame-level fashion, where the surgical phase of each frame serves as the ground truth. As the quality of the extracted spatial features plays a crucial role in the temporal aggregation stage, certain methods used hard-frame detection \cite{RN659} or surgical tools presence supervision \cite{RN655} to enhance the quality of these spatial features. We observe that there is a substantial variation in semantic features among different phases within the limbus region, whereas the regions outside the limbus region exhibit similar appearances (Fig. \ref{fig1}). These observations motivate us to develop a spatial feature extractor that incorporates supervision from both surgical phase and limbus region, with the objective of obtaining limbus region-focused spatial features.

For temporal feature aggregation, early attempts used Long Short-Term Memory (LSTM) or Gated Recurrent Unit (GRU)-based methods \cite{RN660}, \cite{yi2022not} to model the temporal dependencies of spatial features. However, these methods are constrained by their limited temporal receptive field and non-parallel, slow inference. To overcome these limitations, recent studies explored the use of a temporal convolutional network (TCN) or a transformer-based structure, either individually \cite{RN654} or in combination \cite{jin2022trans}. These approaches effectively model long-range temporal relationships, leading to improved accuracy and smoother phase recognition. However, globally aggregating temporal features may neglect important local fine-grained information, such as the dynamic interaction between surgical tools and the eye structure. In specific phases of PCS, such as phacoemulsification, the interaction and movement of surgical tools within a short time window play a more prominent role in distinguishing different surgical phases. These observations serve as a motivation for proposing our long short spatiotemporal aggregation transformer (LS-SAT) network. The design is based on the insight that combining spatial features and local temporal features extracted from neighboring frames can provide fine-grained information, while incorporating global temporal features can offer contextual references for accurate recognition of the current frame.

In this study, we developed a novel phase-specific AR guidance system for PCS. The system recognizes the intraoperative surgical microscopic video using the proposed spatiotemporal learning network, which consists of two stages: a multi-task learning stage for limbus segmentation and spatial feature extraction, and a spatiotemporal aggregation stage for online surgical phase recognition. The segmented limbus region is used for computing guidance parameters and designing AR visual cues. By combining the results of surgical phase recognition, our system offers ophthalmologists a phase-specific AR scene, potentially enhancing their intraoperative skills.

In summary, our major contributions are four-fold:

1) We develop a novel phase-specific AR guidance system, which provides the ophthalmologist with distinct visual cues based on the recognized surgical phase. We evaluate the performance of the developed system on a clinical setup.

2) We design a multi-task learning-based spatial feature extractor for extracting limbus region-focused features, facilitating the computation of essential intraoperative guidance parameters based on the limbus boundary and enhancing the effectiveness of spatiotemporal aggregation.

3) We propose LS-SAT, a transformer-based spatiotemporal learning network that comprehensively uses spatial features, local temporal features, and global temporal features for accurate surgical phase prediction, making it achieve the state-of-the-art results in a publicly available dataset and an in-house dataset.

4) We propose a pipeline for automatically computing the parameters of intraoperative visual cues, which introduces a curvature-based contour points filter to enhance the robustness of limbus ellipse fitting, as well as a regional restricted normal cross-correlation approach for rotation computing.

\section{Related Works}
\label{relatedwork}

\subsection{AR guided microscopic ophthalmic surgery}
Many intraoperative microscopic AR guidance systems \cite{RN653}, \cite{RN662}, and \cite{RN666} have traditionally relied on external optical trackers for tasks like microscope calibration, patient registration, and pose tracking. These systems typically involve overlaying preoperative surgical planning information onto the output video of the microscope or the eyepieces. However, such approaches face incompatibility with ophthalmic surgeries, primarily due to the soft tissue nature of the eyeball, which makes it challenging to affix an optical reference frame to it.

In ophthalmic surgeries, certain studies have explored alternative methods that process online microscope video to design intraoperative guidance systems. For example, \cite{roodaki2015introducing} detected the position and rotation of the surgical instrument tip from the surgical video. These data were then employed to guide real-time Optical Coherence Tomography (OCT) scans. The OCT information was subsequently visualized in the microscopic eyepiece to provide ophthalmologists with depth perception during the procedure. Similarly, \cite{pan2020real} introduced an AR guidance system designed specifically for deep anterior lamellar keratoplasty. Their approach included a deep learning-based method for semantic segmentation and occlusion reconstruction to track the corneal contour. Nevertheless, the AR guidance information in these systems remains somewhat limited in scope and cannot be readily applied to cataract surgery. This is because cataract surgeries involve a complex sequence of surgical steps, each of which necessitates distinct intraoperative guidance information.

There is a limited body of work dedicated to AR guidance specifically tailored for cataract surgery. For example, \cite{RN647} designed a multi-task convolutional neural network (CNN) capable of locating the pupil and classifying the surgical phase. However, the recognized surgical phases were not effectively utilized for intraoperative guidance. \cite{RN648} developed an intraoperative guidance system for cataract surgery, focusing on the placement of IOL. This system functioned by detecting the eye's center and tracking its rotation. Furthermore, \cite{bian2023variation} presented a transformer-enhanced high-resolution network designed for the online recognition of capsulorhexis margins, addressing a specific aspect of cataract surgery. \cite{RN641} employed a deep neural network  to segment surgical instruments and tissue boundaries in real-time, aiming to assist in image-guided ophthalmic surgery. However, a common drawback among these methods is that the presence of overlapping visual information unrelated to phase differentiation can occasionally obstruct the surgeon's field of vision.

\subsection{Online surgical phase recognition}
Early attempts \cite{RN663} utilized CNN-based networks to predict the current surgical phase on a frame-by-frame basis. However, this approach neglected the temporal correlation between frames, leading to inconsistent phase recognition. \cite{wang2022intelligent} employed a 3D-CNN-based framework, which fused sparse and dense channels to recognize the surgical phase in cataract surgery. Nevertheless, this method suffered from slow inference due to the incorporation of 3D CNN. Several studies adopted a two-stage framework. In the first stage, spatial features were extracted using 2D CNN, while in the second stage, these features were aggregated to predict the surgical phase. For example, \cite{RN660} employed the LSTM network, and \cite{yi2022not} used the GRU network to learn the temporal dependencies. However, these methods had limitations in terms of their ability to model long-range temporal dependencies effectively. To improve the network's capacity for long-distance temporal modeling, \cite{RN654} introduced TeCNO, a multi-stage TCN-based network designed to capture global temporal dependencies of spatial features. However, TCN's performance on frames with phase changes was suboptimal due to the use of temporally invariant kernels.

Recent studies have implemented transformer-based architectures for spatiotemporal aggregation. For instance, \cite{RN651} introduced OperA, which utilized attention-regularized transformers for online surgical phase recognition. Additionally, \cite{jin2022trans} presented Trans-SVNet, which obtained temporal embedding features by aggregating spatial features using TCN. They also proposed a transformer-based hybrid embedding model to combine both spatial and temporal features for phase recognition. Furthermore, \cite{RN644} developed an auto-regressive transformer framework that leveraged prediction results from historical frames for recognizing the current phase. \cite{yue2023cascade} proposed a cascade multi-level transformer network for surgical phase recognition, adaptively fusing temporal features from frame-level, temporal features from phase-level, and spatial features. Nevertheless, while long-distance spatiotemporal aggregation contributed to a smooth and stable phase prediction, these methods have been hindered by the absence of local fine-grained features for spatiotemporal aggregation. This limitation has notably impacted their performance, especially in the context of cataract surgeries, where certain phases exhibit high dynamism, and features within a short time window have distinct characteristics.

\begin{figure*}[htbp]
	\centerline{\includegraphics[width=\textwidth]{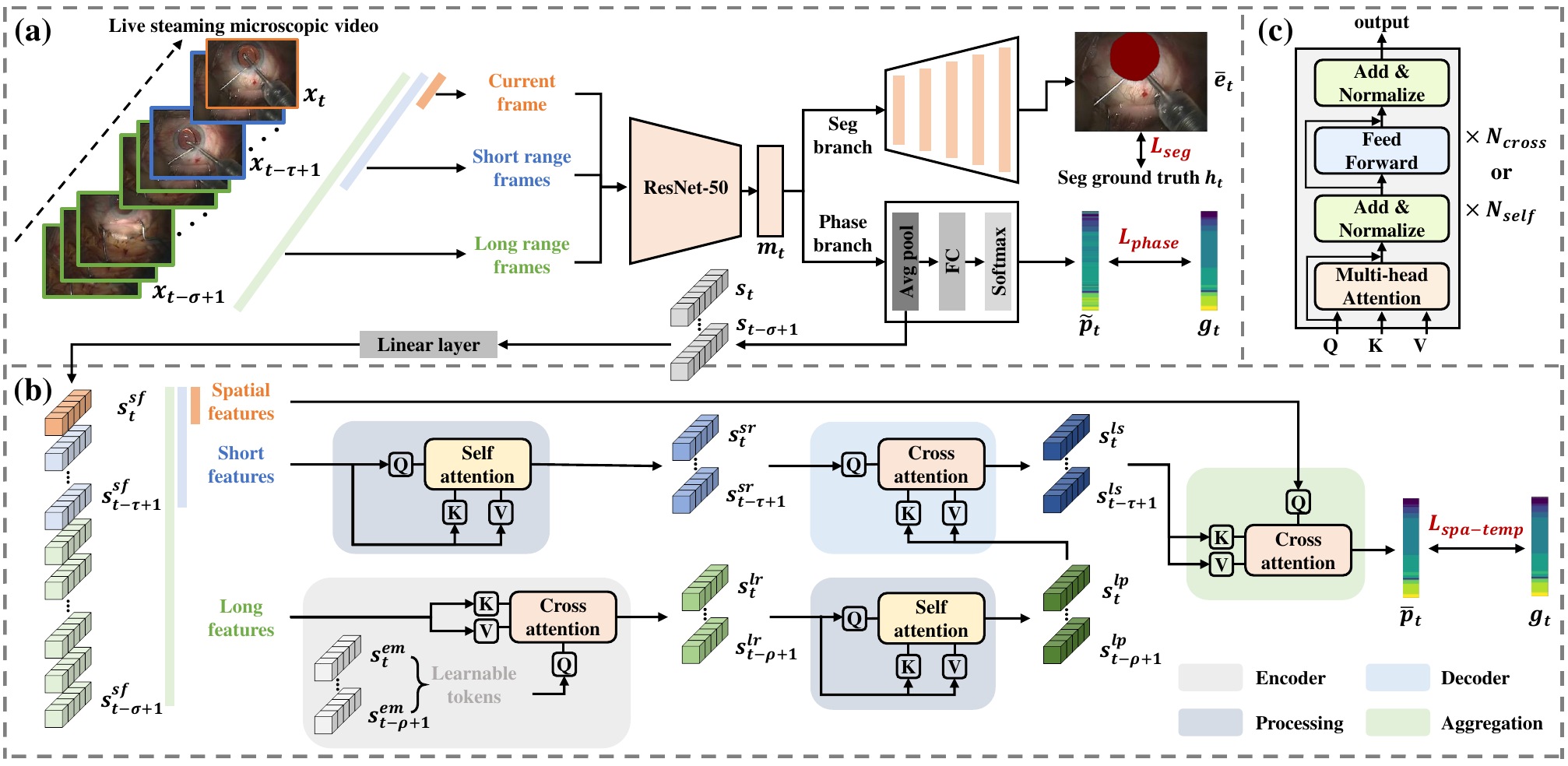}}
	\caption{The architecture of our proposed two-stage spatiotemporal aggregation network. (a)  The spatial feature extraction stage processes live streaming microscopic video frames ${X_t}$ as input and employs multi-task learning to produce the limbus segmentation results ${E_t}$ along with limbus-region-focused spatial features $S_t^{sf}$. (b) The spatiotemporal aggregation stage takes $S_t^{sf}$ as input and employs the proposed SA-SAT network, which integrates an encoding-processing-decoding module and a spatiotemporal aggregation module, to recognize the current surgical phase at time $t$. (c) The structure of the self- and cross-attention module in the SA-SAT network.}
	\label{fig3}
\end{figure*}

\section{Methods}
\subsection{Spatiotemporal network for microscopic surgical video recognition}
\label{sec32}
\subsubsection{Overview}
Fig. \ref{fig3} illustrates the architecture of our proposed spatiotemporal network, comprising two stages: the spatial feature extraction stage and the spatiotemporal aggregation stage. The network takes as input a surgical microscope video denoted as ${X_t} = \{ {{x}_{t -\sigma +1}}, \cdots ,{x_t}\}$, where each frame ${x_i} \in {\mathbb{R}^{H \times W \times C}}$,  $H$ and $W$ represent the height and width, and $C = 3$ denotes the number of channels. Note that $1 \le t \le T$, and $T$  represents the video length. The spatial feature extraction stage (Fig. \ref{fig3} (a)) is designed as a multi-task learning structure. It maps ${X_t}$ to a set of spatial features denoted as ${S_t} = \{ {{s}_{t -\sigma +1}}, \cdots ,{s_t}\} $, where each ${s_i} \in {\mathbb{R}^{C_m}}$ and $C_m=2048$ represents the number of channels. The stage also generates a set of segmented limbus regions represented as ${E_t} = \{{{\bar e}_{t -\sigma +1 }}, \cdots ,{\bar e_t}\} $, where each ${\bar e_i} \in {\mathbb{R}^{H \times W \times {C_0}}}$ and ${C_o} = 2$ represents the number of output channels. The spatiotemporal aggregation stage (Fig. \ref{fig3} (b)) firstly employs a linear layer to reduce the dimension of ${S_t}$ by a factor of $\kappa $, resulting in $S_t^{sf} = \{ s_{t -\sigma +1 }^{sf}, \cdots ,s_t^{sf}\} $, where each  $s_i^{sf} \in {\mathbb{R}^{C_m/\kappa }}$. 
The extracted spatial features $S_t^{sf}$ are then passed through the LS-SAT model. This model incorporates an  encoding-processing-decoding framework for the fusion of long-short temporal features and an aggregation module for the fusion of spatiotemporal features. This process yields the surgical phase probability ${\bar p_t} \in {\mathbb{R}^{{K_s}}}$, where ${K_s}$ represents the number of surgical phases.

\subsubsection{Limbus region-focused spatial feature extraction} 
We designed a multi-task learning structure to jointly train the tasks of limbus region segmentation and spatial feature extraction, instead of training them separately, for two reasons: 1) During PCS, the appearances and surgical tools within the limbus region exhibit significant variations across different surgical phases, while other regions like the sclera tend to have similar semantic features (Fig. \ref{fig1}). By training these two tasks together, the model can effectively capture and leverage the shared semantic information and correlations between them. This enables the model to generate spatial features that are specifically focused on the limbus region, enhancing comprehension of the surgical scene. 2) By sharing the feature extraction backbone parameters between the two tasks, we promote computational efficiency, which is crucial for achieving real-time intraoperative guidance.

Our multi-task-based spatial feature extraction network utilizes ResNet-50 \cite{RN665} as the backbone, employing hard parameter sharing. For each input frame  ${x_t}$, the backbone generates an output feature map ${m_t} \in {\mathbb{R}^{{H_m} \times {W_m} \times {C_m}}}$, where ${H_m}$ and ${W_m}$ represent the height and width. The feature map   is then fed into two branches: the frame-wise phase recognition branch and limbus segmentation branch. The frame-wise phase recognition branch comprises an average pooling layer, a fully connected layer, and a softmax layer. The output of this branch is denoted as ${\tilde p_t} \in {\mathbb{R}^{{K_s}}}$, representing the phase probability of ${x_t}$. The limbus segmentation branch incorporates a decoder with upsampling and concatenation, resembling the U-net \cite{RN668} architecture. The output of this branch is ${\bar e_t} \in {\mathbb{R}^{H \times W \times {C_0}}}$.

After training the spatial feature extraction stage, we extract the spatial feature ${s_t}$ for each input frame ${x_t}$ by taking the output of the average pooling layer in the frame-wise phase recognition branch. These spatial features ${S_t}$ 
are dimensional reduced with a linear layer to produce limbus region-focused spatial features $S_t^{sf}$, which serve as the input for the subsequent spatiotemporal aggregation stage. 

\subsubsection{LS-SAT-based surgical phase recognition} 
Our LS-SAT network effectively captures both long-range and local temporal dependencies of the input spatial features. It further aggregates the extracted limbus-region focused spatial features to facilitate online phase recognition. Our network is built upon the transformer architecture \cite{RN664}, which leverages the attention mechanism (Fig. \ref{fig3} (c)) to model long-range spatial and temporal interactions in a parallel manner, making it well-suited for surgical video recognition.

We employ an encoding-processing-decoding framework to capture the interplay between short and long-range features. When dealing with long-range features, one natural approach for encoding involves the use of a self-attention-based transformer encoder. However, this approach exhibits a time complexity of $O({{\sigma }^{2}}\times {{C}_{m}}/\kappa )$, which grows quadratically with the sequence length of the long features, thereby constraining its efficacy in modeling long-range temporal dependencies. To address this challenge, we opt for a memory compression-based encoder \cite{xu2021long} to enhance efficiency. Specifically, we introduce learnable tokens ${{S}^{em}}=\{s_{t-\rho+1}^{em},\cdots ,s_{t}^{em}\}$  and incorporate them, along with the complete long features $S_{t-\sigma +1:t}^{sf} \in {\mathbb{R}^{\sigma \times (C_m/\kappa )}}$, into a cross-attention module. Here, the learnable tokens function as Query (Q), while the full long features operate as Key (K) and Value (V). This operation can be  expressed as
\begin{equation}
	{EC({{S}^{em}},S_{t-\sigma +1:t}^{sf})=softmax (\frac{{{S}^{em}}\cdot S{{_{t -\sigma+1 :t}^{sf}}^{T}}}{\sqrt{{{C}_{m}}/\kappa }})S_{t -\sigma +1:t}^{sf}}.
\end{equation}

Notably, due to the fact that $\rho\ll\sigma$, this encoder exhibits a substantially reduced time complexity of $O(\rho \times \sigma \times {{C}_{m}}/\kappa )$, rendering it much more efficient compared to the self-attention-based encoder. We denote the number of cross-attention layers as ${N_{cross}^{ec}}$, and the output of the encoder as $S_{\rho }^{lr}=\{s_{t-\rho+1}^{lr},\cdots ,s_{t}^{lr}\}$. $S_{\rho }^{lr}$ is directed into the processing module, where the multi-head self-attention module utilizes it as the Q, K, and V. This operation can be expressed as
\begin{equation}
	{PC}({S_{\rho }^{lr}}) = softmax(\frac{{{S_{\rho }^{lr}} \cdot {{{S_{\rho }^{lr}}}^{T}}}}{{\sqrt {{{C}_{m}}/\kappa} }}){S_{\rho }^{lr}}.
\end{equation}

The output of the processing module is denoted as $S_{\rho }^{lp}=\{s_{t-\rho+1}^{lp},\cdots ,s_{t}^{lp}\}$, and the number of self-attention layers is represented by ${N_{self}^{lr}}$.

To capture short-range dependencies, we concentrate on a specific window of $\tau $ past frames at time $t$, which is expressed as $S_{t - \tau+1:t}^{sf} \in {\mathbb{R}^{\tau \times (C_m/\kappa )}}$. For modeling temporal dependencies within this window, we employ another processing module. The output of this module is denoted as ${{S}^{sr}}=\{s_{t - \tau+1}^{sr},\cdots ,s_{t}^{sr}\}$ and the number of self-attention layers is represented by ${N_{self}^{sr}}$.

We aggregate short and long temporal features using a cross-attention-based decoder module that takes Q and K from different sources. Here, we take the short-range temporal features ${{S}^{sr}}$ as Q and the long-range compressed temporal features ${{S}^{lp}}$ as K and V. The encoder module can be formulated as
\begin{equation}
	{DC}({{S}^{sr}}, {{S}^{lp}}) = softmax(\frac{{{{S}^{sr}} \cdot {{{S}^{lp}}^{T}}}}{{\sqrt {{{C}_{m}}/\kappa} }}){{S}^{lp}}.
\end{equation}

We denote the number of cross-attention layers as ${N_{cross}^{dc}}$, and the output of the decoder module as long-short temporal features ${{S}^{ls}}=\{s_{t-\tau+1}^{ls},\cdots ,s_{t}^{ls}\}$.

We then aggregate the spatiotemporal features using another one-layer cross-attention-based aggregation module. We take $s_t^{sf}$ as Q and ${{S}^{ls}}$ as K and V, promoting the interaction between the limbus-region focused spatial features and the long-short temporal features. The aggregation module can be formulated as
\begin{equation}
	{AG}(s_t^{sf}, {{S}^{ls}}) = softmax(\frac{{s_t^{sf} \cdot {{{S}^{ls}}^{T}}}}{{\sqrt {{{C}_{m}}/\kappa} }}){{S}^{ls}}.
\end{equation}

The output of the aggregation module is denoted as ${{\hat{p}}_{t}}\in {{\mathbb{R}}^{C_m/\kappa }}$, which is connected to a fully connected layer, and a softmax layer to produce ${\bar p_t}$ as
\begin{equation}
	{{\bar{p}}_{t}}=softmax({{W}_{p}}\cdot {{\hat{p}}_{t}}),
\end{equation}
where ${W_p} \in {\mathbb{R}^{(C_m/\kappa ) \times {K_s}}}$.

\begin{figure*}[htbp]
	\centerline{\includegraphics[width=\textwidth]{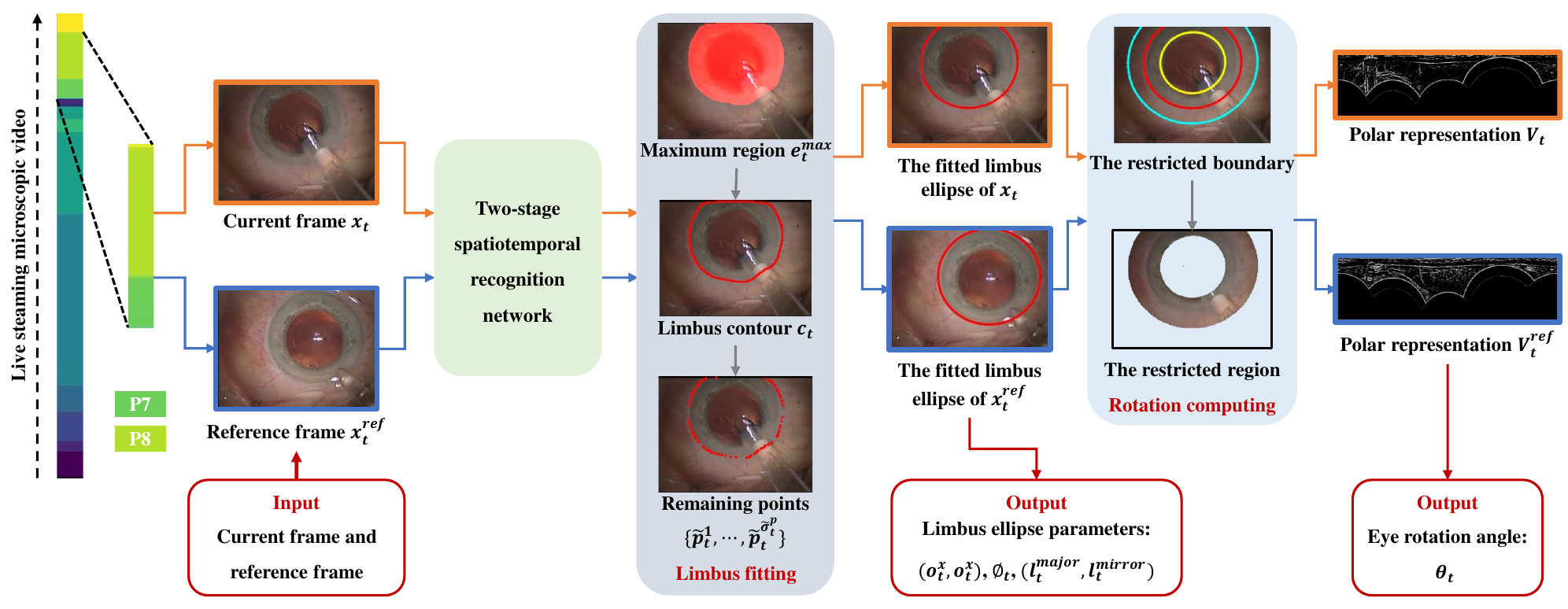}}
	\caption{Pipeline for intraoperative guidance parameter computation: using P7 (lens implant) as an example, the first frame at P7 serves as the reference frame. We input this reference frame along with the current frame into our proposed two-stage shape correction network for phase recognition and limbus segmentation. Initially, we fit the limbus with an ellipse and generate the ellipse parameters. Subsequently, we focus on a restricted region and derive the polar representation for both the reference frame and the current frame. After this, we compute and produce the eye rotation angle using the normalized cross-correlation-based method. Note that at each stage, the limbus region and polar representation of the reference frame are calculated once, while those of the current frame are computed every time.}
	\label{fig4}
\end{figure*}

\subsubsection{Training} 
We define the ground truth for the surgical phase as ${g_t} \in {\mathbb{R}^{{K_s}}}$, and the ground truth for the limbus segmentation as  ${h_t} \in {\mathbb{R}^{H \times W \times {C_o}}}$. The spatial feature extraction stage is trained to extract spatial features, which are then utilized to train the spatiotemporal aggregation stage.

The loss function for training the spatial feature extraction stage consists of two parts: the phase recognition part ${L_{phase}}$ and the limbus segmentation part ${L_{seg}}$. We employ cross-entropy loss for ${L_{phase}}$, given by
\begin{equation}
	{L_{phase}} =  - \frac{1}{{{K_s}}}\sum\limits_{s = 1}^{{K_s}} {{g_{t,s}}} \log {\tilde p_{t,s}}.
\end{equation}

For ${L_{seg}}$, we utilize a hybrid loss of cross-entropy and Dice, expressed as
\begin{equation}
	\begin{split}
		{L_{seg}} =  - \frac{1}{{H \times W \times {C_0}}}\sum\limits_{c = 1}^{{C_0}} {\sum\limits_{i = 1}^{H \times W} {\bar e_{t,i}^c} } \log h_{t,i}^c \\
		+ \alpha (1 - \frac{{2\sum\limits_{c = 1}^{{C_0}} {\sum\limits_{i = 1}^{H \times W} {\bar e_{t,i}^ch_{t,i}^c} } }}{{\sum\limits_{c = 1}^{{C_0}} {\sum\limits_{i = 1}^{H \times W} {\bar e_{t,i}^c}  + \sum\limits_{c = 1}^{{C_0}} {\sum\limits_{i = 1}^{H \times W} {h_{t,i}^c} } } }}),
	\end{split}
\end{equation}
where $\alpha $ is a weighting coefficient. The loss function of the spatial feature extraction stage is represented as
\begin{equation}
	{L_{sf}} = {L_{phase}} + \beta {L_{seg}},
\end{equation}
where $\beta $ is a weighting coefficient. 

Due to significant variations in the duration of each surgical step within the dataset, a class-imbalanced problem arises. To address this, we employ a weighted cross-entropy loss during the training of the spatiotemporal aggregation stage, denoted as
\begin{equation}
	{L_{spa - temp}} =  - \frac{1}{{{K_s}}}\sum\limits_{s = 1}^{{K_s}} {{w_s}{g_{t,s}}} \log {\bar p_{t,s}},
\end{equation}
where ${w_s}$ represents the weight parameter, which is inversely proportional to the phase frequencies \cite{RN667}.

\subsection{Intraoperative guidance parameters computation}
\label{sec33}
The spatiotemporal model is deployed intraoperatively to recognize the surgical phase ${p_t}$ and obtain the limbus segmentation result ${e_t}$. The boundary of the segmented limbs region can be fitted as an ellipse \cite{RN648}\cite{RN643}. In this study, we leverage the segmented limbus region ${e_t}$ at a given time $t$ to compute two types of essential intraoperative guidance parameters: 1) the fitted limbus ellipse parameters, including the central coordinates $(o_t^x,o_t^y)$, the lengths of the major and minor axes $(l_t^{major},l_t^{minor})$, and the rotation angle of the fitted ellipse ${\phi _t}$; 2) the eye rotation angle ${\theta _t}$, which represents the rotational displacement of the current frame relative to a reference frame. 

Based on clinical experience, the parameters of specific intraoperative guidance visual cues, such as the position of the incision curve and the range of the capsulorhexis, can be determined using the fitted ellipse parameters. Additionally, the eye rotation parameters can be utilized to design rotation reference visual cues, which play a crucial role in achieving accurate intraoperative IOL alignment. The computation pipeline of these intraoperative parameters is demonstrated in Fig. \ref{fig4}, and the following sections elaborate on the details of this pipeline.

\subsubsection{Fitted limbus ellipse parameters computation} 
We compute the fitted limbus ellipse parameters at all surgical phases. Firstly, we remove potential mis-segmented regions in ${e_t}$ by extracting the maximum connected region, denoted as $e_t^{\max } = \max \{ e_t^1, \cdots ,e_t^{\sigma _t^e}\} $, where $\sigma _t^e$ represents the number of regions. We then extract the contour of $e_t^{\max }$, denoted as ${c_t}$, to obtain a set of contour points $\{ p_t^1, \cdots ,p_t^{\sigma _t^p}\} $, where $\sigma _t^p$ represents the number of points. The curvature of each contour point $p_t^i$ can be computed using its neighboring points $p_t^{i - 1}$ and $p_t^{i + 1}$, represented as $\ell _t^i = curv\{ p_t^{i - 1},p_t^i,p_t^{i + 1}\}$. Next, we define a curvature threshold of ${\mu _{curv}}$, and exclude the  $i - th$ boundary point or outlier when $\ell _t^i < {\mu _{curv}}$. The remaining boundary points are denoted as $\{ \tilde p_t^1, \cdots ,\tilde p_t^{\tilde \sigma _t^p}\} $, which can be utilized for fitting the limbus ellipse. The limbus ellipse parametric equation is given by
\begin{equation}
	\left\{ {\begin{array}{*{20}{c}}
			{{x_t} = o_t^x + l_t^{major}\cos (\varphi )\cos ({\phi _t}) - l_t^{minor}\sin (\varphi )\sin ({\phi _t})}\\
			{{y_t} = o_t^y + l_t^{major}\cos (\varphi )\sin ({\phi _t}) + l_t^{minor}\sin (\varphi )\cos ({\phi _t})}
	\end{array}} \right..
\end{equation}

We construct an error function representing the sum of squared distances from each remaining boundary point to the fitted ellipse, denoted as
\begin{equation}
	{E_x} = \sum\limits_{i = 1}^{\tilde \sigma _t^p} {((} x_t^i-{x_t}{)^2} + {(y_t^i-{y_t})^2}).
\end{equation}

Finally, the limbus ellipse parameters, including $(o_t^x,o_t^y)$, $(l_t^{major},l_t^{minor})$, and ${\phi _t}$ can be determined through optimization of the nonlinear least-squares problem utilizing the Levenberg-Marquardt algorithm. In this optimization process, $(o_t^x,o_t^y)$ is initially set to the average of all boundary points, $(l_t^{major},l_t^{minor})$ is initialized at half the average distance from all boundary points to $(o_t^x,o_t^y)$, and ${\phi _t}$ is initialized to 0.

\begin{table}[!t]
	\centering
	\caption{The definition of our designed AR visual cues.}
	\begin{tabular}{p{8em}<{\centering}p{20em}}
		\toprule
		\multicolumn{1}{c}{\textbf{AR visual cue}} & \multicolumn{1}{c}{\textbf{Definition}} \\
		\midrule
		\multicolumn{1}{m{2.4cm}} {Fitted limbus contour (FLC)} & \multicolumn{1}{m{5.4cm}}{The fitted limbus ellipse with central coordinates $(o_{t}^{x},o_{t}^{y})$, the lengths of the major and minor axes $(l_{t}^{major},l_{t}^{minor})$, and a rotation angle ${{\phi }_{t}}$.} \\ \hline
		
		\multicolumn{1}{m{2.4cm}} {Primary incision guideline (PIG)} & \multicolumn{1}{m{5.4cm}}{A line starts from $(o_{t}^{x},o_{t}^{y})$, extending $(l_{t}^{major}+l_{t}^{minor})\times 0.3$, and forming an included angle of 95° with RRL.} \\ \hline
		
		\multicolumn{1}{m{2.4cm}} {Primary incision curve (PIC)} & \multicolumn{1}{m{5.4cm}}{A curve within FPC, having a length equal to the maximum axial distance of the knife, and with PIG serving as the dividing line.} \\ \hline
		
		\multicolumn{1}{m{2.4cm}} {Secondary incision guideline (SIG)} & \multicolumn{1}{m{5.4cm}}{A line starts from $(o_{t}^{x},o_{t}^{y})$, extending $(l_{t}^{major}+l_{t}^{minor})\times 0.3$, and forming an included angle of 175° with RRL.} \\ \hline
		
		\multicolumn{1}{m{2.4cm}} {Secondary incision curve (SIC)} & \multicolumn{1}{m{5.4cm}}{A curve within FPC, having a length equal to the maximum axial distance of the secondary incision knife, and with the SIG serving as the dividing line.} \\ \hline
		
		\multicolumn{1}{m{2.4cm}} {Capsulorhexis circular range (CCR)} & \multicolumn{1}{m{5.4cm}}{A circle with a diameter of $(l_{t}^{major}+l_{t}^{minor})/2$.} \\ \hline
		
		\multicolumn{1}{m{2.4cm}} {Rotation reference line (RRL)} & \multicolumn{1}{m{5.4cm}}{A line passes through $(o_{t}^{x},o_{t}^{y})$, forming an included angle of ${{\theta }_{t}}$ with the horizontal direction, and having a length of $l_{t}^{major}\times 1.2$.} \\
		\bottomrule
	\end{tabular}%
	\label{table1}%
\end{table}%

\subsubsection{Eye rotation parameters computation} 
We compute the eye rotation angle at P7 (lens implant) and P8 (VA removal) but exclude other phases because there is no requirement to display the rotation reference line during those phases. For current frame ${x_t}$ with a predicted surgical phase ${p_t}$, we take the first frame of phase ${p_t}$ as the reference frame $x_t^{ref}$. Our eye rotation method aims to compute the rotation degree ${\theta _t}$ between ${x_t}$ and $x_t^{ref}$.

We observe that regions around the limbus boundary, such as iris and conjunctival vessels, display distinguishable texture features (Fig. \ref{fig1} (c)). Therefore, our initial step is to limit the computation of rotation near the limbus boundary. We define the restricted region using the parameters $d_t^{in} = (l_t^{major} + l_t^{minor})/2/\lambda _t^{in}$ inside the limbus boundary and $d_t^{out} = (l_t^{major} + l_t^{minor})/2/\lambda _t^{{\rm{out}}}$ outside the limbus boundary, where $\lambda _t^{in}$ and $\lambda _t^{out}$ represent the scale factors. The restricted region can be expressed as
\begin{equation}
	\begin{aligned}
		{U_t} &= \{ ({x_t},{y_t})| - d_t^{in} \le \\
		\sqrt{{{({{x}_{t}}-o_{t}^{x})}^{2}}+{{({{y}_{t}}-o_{t}^{y})}^{2}})} & - (l_t^{major} + l_t^{minor})/2 \le d_t^{out}\}.
	\end{aligned}
\end{equation}

Next, we proceed to convert the $t - th$ restricted region ${U_t}$ and the reference restricted region $U_t^{ref}$ into polar coordinate representations, denoted as ${V_t}$ and $V_t^{ref}$, respectively. The rotation between ${U_t}$ and $U_t^{ref}$ can then be equivalently represented as the displacement between ${V_t}$ and $V_t^{ref}$. 

To achieve robust displacement estimation, we employ the normalized cross-correlation-based method. The normalized cross-correlation between ${V_t}$ and $V_t^{ref}$ is given by
\begin{equation}
	\begin{aligned}
		D(u,v) = &\sum\limits_{i = 1}^{{N_{cc}}} {(V_t^{ref}({x_i},{y_i}) - \mu (V_t^{ref}))*} \\
		&{({V_t}({x_i} - u,{y_i} - v) - \mu ({V_t}))/(\sigma _t^{ref}*{\sigma _t})},
	\end{aligned}
\end{equation}
where $\mu (V_t^{ref})$ and $\mu ({V_t})$ represent the mean of $V_t^{ref}$ and ${V_t}$. Similarly, $\sigma _t^{ref}$ and ${\sigma _t}$ represent the standard deviation of $V_t^{ref}$ and ${V_t}$, respectively. ${N_{cc}}$ represents the number of pixels in ${V_t}$. By solving $\max D(u,v)$, we obtain the displacement $({u_{\max }},{v_{\max }})$. Subsequently, the displacement is converted into angular representation to obtain the rotation degree ${\theta _t}$.

\begin{figure}[!t]
	\centerline{\includegraphics[width=\columnwidth]{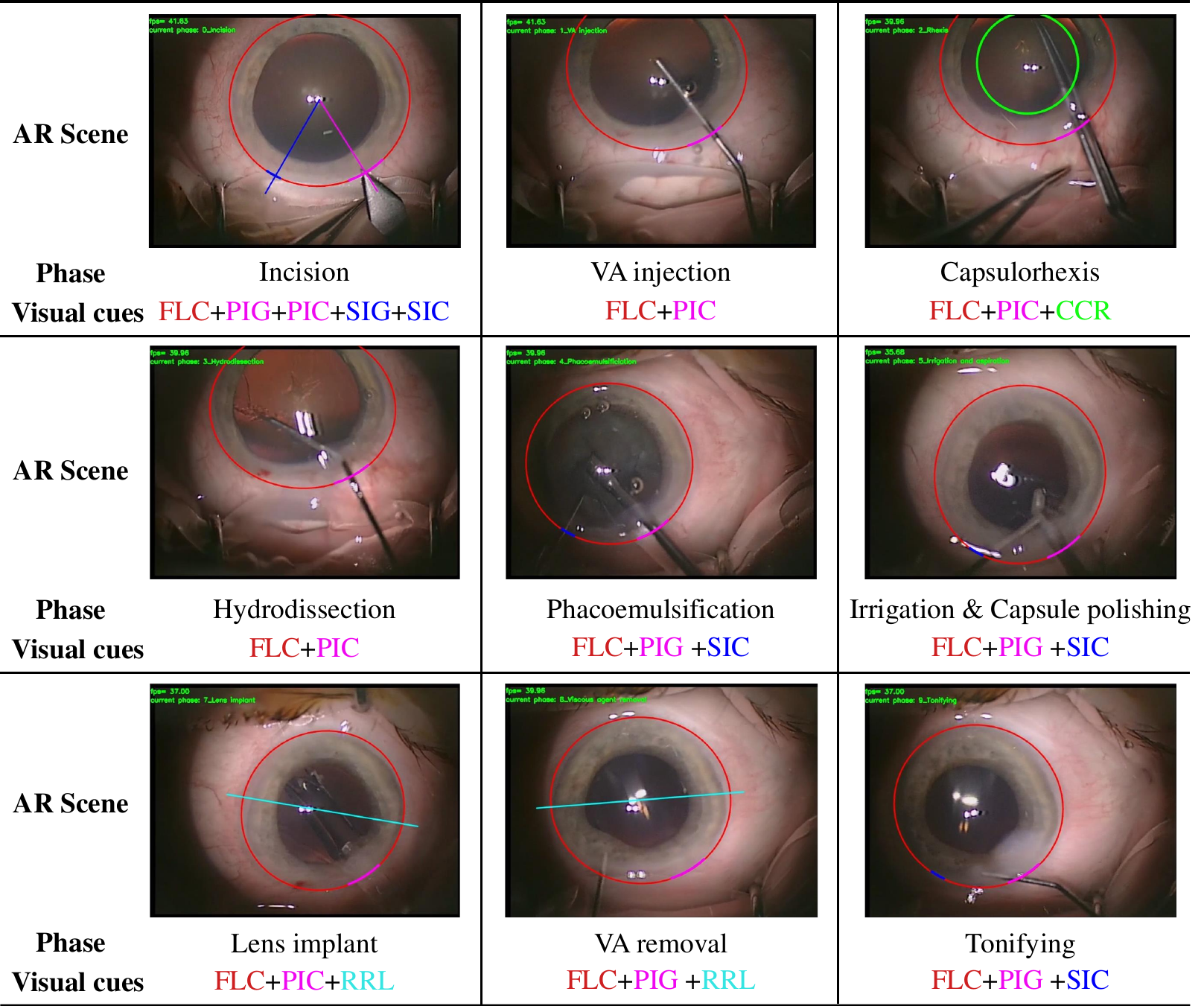}}
	\caption{Different AR visual cues are represented using varying colors, and the same color scheme is maintained for each visual cue across different surgical phases.}
	\label{fig5}
\end{figure}

\subsection{Phase-specific AR Guidance in PCS}
\label{sec34}
We utilize the intraoperative guidance parameters to design AR visual cues, which are then combined with surgical phase information to provide a phase-specific AR scene for ophthalmologists.

By observing the attention behavior of ophthalmologists during PCS, we have designed seven visual cues in collaboration with ophthalmologists, and their definitions are listed in Table \ref{table1}. Following \cite{RN661}, we divide the PCS into ten phases: incision, viscous agent (VA) injection, capsulorhexis, hydrodissection, phacoemulsification, irrigation, capsule polishing, lens implant, VA removal, and tonifying. Note that irrigation and capsule polishing in our XH-CaTa dataset are merged into a single surgical phase as they share the same combination of visual cues. We present the AR scene with different combinations of visual cues for each surgical phase in Fig. \ref{fig5}. 

\section{Experiments and results}
\subsection{Dataset}
Our methods were evaluated on two datasets: 1) Cataract-101 \cite{RN661}, a publicly available dataset for phase recognition in cataract surgery, and 2) XH-CaTa, an in-house dataset from an in-house dataset from the department of ophthalmology, Xinhua Hospital Affiliated to Shanghai Jiao Tong University School of Medicine.

Cataract-101 consists of 101 videos with a frame rate of 25 frames per second (fps), and the total duration of the dataset is around 14 hours. Each video was subsampled to 1 fps, and each frame has a resolution of $720\times 540$ pixels. The dataset was annotated into 10 surgical phases. Following \cite{RN650}, we divided the dataset into 73 cases for training and 28 cases for testing, with an additional 13 cases from the training set reserved for validation.

XH-CaTa comprises 43 videos with a frame rate of 30 fps, and the total duration of the dataset is around 8.6 hours. Similar to Cataract-101, each video in XH-CaTa was subsampled to 1 fps, and each frame has a resolution of $1920\times 1080$ pixels. XH-CaTa was annotated into 9 surgical phases (irrigation and capsule polishing are considered the same phase) by two experienced ophthalmologists. We used 30 cases for training and 13 cases for testing, with 8 cases from the training set used for validation.

\begin{figure}[!t]
	\centerline{\includegraphics[width=\columnwidth]{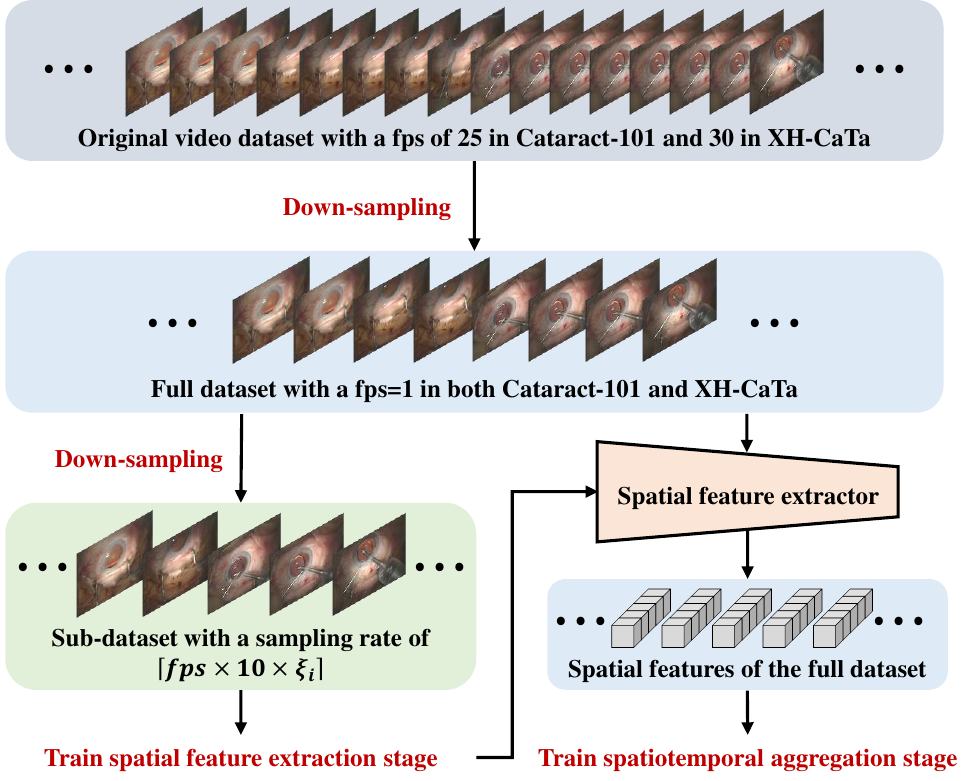}}
	\caption{Dataset processing workflow. Initially, we down-sample the original video to create the full dataset, and this is where surgical phase annotation is performed. Subsequently, we down-sample the full dataset to obtain the sub-dataset, which is used for annotating limbus regions. We train the spatial feature extraction stage using the sub-dataset and then employ the spatial feature extractor to extract spatial features for the full dataset. Finally, these spatial features are utilized to train the spatiotemporal aggregation stage. }
	\label{fig10}
\end{figure}

The workflow of our dataset processing is shown in Fig. \ref{fig10}. To facilitate limbus region annotation, we employed a larger subsampling rate for both Cataract-101 and XH-CaTa datasets, resulting in a sparser sub-dataset. This approach effectively reduced the annotation workload. The subsampling rate is defined as $\left\lceil fps\times 10\times {{\xi }_{i}} \right\rceil $, where ${{\xi }_{i}}$ represents the normalized reciprocal frequency of the $i-th$ surgical phase. By applying subsampling, each surgical phase in the sub-dataset has an equal number of frames, thereby avoiding the issue of class imbalance during the training of the spatial feature extraction stage. The limbus regions in the sub-dataset were manually delineated by two non-M.D. experts. 

\subsection{Implementation Details}
We implemented our method in PyTorch, using two NVIDIA GeForce GTX 3090 GPUs. The backbone ResNet-50 \cite{RN665} was pre-trained on the ImageNet  \cite{RN669} dataset. Each frame was resized to a resolution of $256\times 256$ pixels. Data augmentation strategies including color jitter, random crop and random rotation were applied. The spatial feature extraction stage was trained using the Adam optimizer for 50 epochs. The batch size was set to 256, and the learning rate was set to 5e-5 for the backbone and 5e-4 for the segmentation branch and the fully connected layer. The spatiotemporal aggregation stage was trained using the Adam optimizer for 30 epochs, with a batch size of 1 and a learning rate of 1e-4.

For hyper-parameters, after fine-tuning, we set the dimensional reduction factor $\kappa =16$, the length of the short window $\tau =10$, the length of the long window $\sigma =120$, the number of transformer layers ${N_{cross}^{ec}={N_{cross}^{dc}}=1}$, ${N_{self}^{lr}}={N_{self}^{sr}}=4$, the weighting coefficient $\alpha \text{=}0.6$ and $\beta \text{=}0.5$, the curvature threshold ${{\mu }_{curv}}=0.7$, and the scale factors $\lambda _{t}^{in}=\lambda _{t}^{out}=3$.

\begin{table*}[t]
	\centering
	\caption{Comparison results for phase recognition on Cataract-101 and XH-CaTa.}
	\begin{tabular}{p{12.40em}|p{3.60em}<{\centering}p{3.60em}<{\centering}p{3.60em}<{\centering}p{3.60em}<{\centering}|p{3.60em}<{\centering}p{3.60em}<{\centering}p{3.60em}<{\centering}p{3.60em}<{\centering}}
		\toprule
		\multirow{2}[4]{*}{\textbf{Method}} & \multicolumn{4}{p{14.4em}<{\centering}|}{\textbf{Cataract-101}} & \multicolumn{4}{p{14.4em}<{\centering}}{\textbf{XH-CaTa}} \\
		\cmidrule{2-9}    \multicolumn{1}{c|}{} & \textbf{Acc (\%)} & \textbf{Pre (\%)} & \textbf{Rec(\%)} & \textbf{Jac (\%)} & \textbf{Acc (\%)} & \textbf{Pre (\%)} & \textbf{Rec (\%)} & \textbf{Jac (\%)} \\
		\midrule
		ResNet-50 \cite{RN665} & 77.7±6.3 & 75.8±10.3 & 74.7±9.7 & 61.3±8.4 & 75.2±7.5 & 75.4±7.6 & 70.4±9.7 & 57.8±12.1 \\
		SV-RCNet \cite{RN660} & 85.8±5.0 & 83.0±9.2 & 83.1±8.5 & 71.1±11.2 & 83.5±5.0 & 80.6±6.4 & 77.4±9.8 & 66.1±8.3 \\
		TMRNet \cite{RN649} & 87.3±7.5 & 84.9±8.3 & 83.1±9.4 & 72.6±6.1 & 84.0±8.3 & 77.6±8.8 & 76.2±7.4 & 64.5±10.7 \\
		TeCNO \cite{RN654} & 89.2±7.4 & 86.6±8.7 & 86.2±6.3 & 76.4±8.6 & 86.3±7.6 & 82.9±7.7 & 82.6±6.9 & 71.7±9.8 \\
		OperA \cite{RN651} & 90.3±4.8 & 86.8±5.9 & 85.8±6.9 & 76.8±9.8 & 87.7±4.2 & 85.1±5.3 & 84.8±6.4 & 74.2±7.5 \\
		Trans-SVNet \cite{jin2022trans} & 91.9±6.9 & 89.1±4.6 & 88.4±5.3 & 80.5±7.7 & 90.1±4.9 & 87.5±5.0 & 85.3±7.0 & 76.5±9.3 \\
		LS-SAT (ours) & \textbf{93.6±4.3} & \textbf{91.7±5.1} & \textbf{91.1±4.6} & \textbf{84.5±7.3} & \textbf{92.4±4.1} & \textbf{89.5±5.4} & \textbf{89.2±5.1} & \textbf{81.5±5.8} \\
		\bottomrule
	\end{tabular}%
	\begin{tablenotes}
		\footnotesize
		\item The best results are highlighted with bold font.
	\end{tablenotes}
	\label{table2}%
\end{table*}%

\begin{table*}[t]
	\centering
	\caption{Results of different spatial feature extractor on Cataract-101 dataset.}
	\begin{tabular}{p{3.0em}<{\centering}p{9.8em}<{\centering}p{4.8em}<{\centering}p{4.8em}<{\centering}p{4.8em}<{\centering}p{4.8em}<{\centering}p{4.8em}<{\centering}}
		\toprule
		\multicolumn{1}{c}{\textbf{Stage}} & \textbf{Method} & \textbf{Acc (\%)} & \textbf{Pre (\%)} & \textbf{Rec (\%}) & \textbf{Jac (\%}) & \textbf{Dice (\%)} \\
		\midrule
		\multicolumn{1}{c}{\multirow{3}[2]{*}{Stage I}} & Phase only & 77.7±6.3 & 75.8±10.3 & 74.7±9.7 & 61.3±8.4 & — \\
		& Segmentation only & —     & —     & —     & —     & 93.0±3.3 \\
		& Multi-task & 81.9±7.2 & 79.7±9.5 & 74.5±8.7 & 63.1±7.9 & \textbf{94.8±2.8} \\
		\midrule
		\multicolumn{1}{c}{\multirow{2}[2]{*}{Stage II}} & w/o limbus-focused & 90.1±6.4 & 86.6±5.6 & 85.5±5.2 & 76.6±6.8 & — \\
		& with limbus-focused & \textbf{93.6±4.3} & \textbf{91.7±5.1} & \textbf{91.1±4.6} & \textbf{84.5±7.3} & — \\
		\bottomrule
	\end{tabular}%
	
	\begin{tablenotes}
		\footnotesize
		\item Stage I: the spatial feature extraction stage; Stage II: the spatiotemporal aggregation stage.
		\item The best results are highlighted with bold font.
	\end{tablenotes}
	\label{table3}%
\end{table*}%

\begin{table}[t]
	\centering
	\caption{Results of different feature combination on Cataract-101 dataset.}
	\begin{tabular}{p{1.3em}<{\centering}p{1.3em}<{\centering}p{1.3em}<{\centering}|p{3em}<{\centering}p{3em}<{\centering}p{3em}<{\centering}p{3em}<{\centering}}
		\toprule
		\textbf{LF} & \textbf{SF} & \textbf{SF} & \textbf{Acc (\%)} & \textbf{Pre (\%)} & \textbf{Rec (\%)} & \textbf{Jac (\%)} \\
		\midrule
		\checkmark     & \multicolumn{1}{c}{} & \multicolumn{1}{c|}{} & 91.1±5.9 & 88.0±7.7 & 87.2±7.1 & 78.8±8.1 \\
		\multicolumn{1}{c}{} & \checkmark     & \multicolumn{1}{c|}{} & 90.1±6.4 & 86.6±7.4 & 85.5±6.0 & 76.6±8.8 \\
		\checkmark     & \checkmark     & \multicolumn{1}{c|}{} & 93.1±4.5 & 90.3±5.8 & 89.8±4.9 & 82.8±6.7 \\
		\multicolumn{1}{c}{} & \checkmark     & \checkmark     & 92.1±5.0 & 88.8±6.3 & 88.1±6.1 & 80.4±7.6 \\
		\checkmark     & \multicolumn{1}{c}{} & \checkmark     & 91.9±6.2 & 89.1±6.6 & 88.4±5.2 & 80.5±8.2 \\
		\checkmark     & \checkmark     & \checkmark     & \textbf{93.6±4.3} & \textbf{91.7±5.1} & \textbf{91.1±4.6} & \textbf{84.5±7.3} \\
		\bottomrule
	\end{tabular}%
	\begin{tablenotes}
		\footnotesize
		\item LF: long feature; SF: short feature; SF: spatial feature.
		\item The best results are highlighted with bold font.
	\end{tablenotes}
	\label{table4}%
\end{table}%

\subsection{Evaluation of the spatiotemporal aggregation network}
\subsubsection{Comparisons with state-of-the-art methods}
We compared our LS-SAT network with several state-of-the-art methods in online surgical phase recognition, including 1) ResNet-50 \cite{RN665}, which serves as the backbone of our spatial feature extractor; 2) SV-RCNet \cite{RN660}, a LSTM-based temporal aggregation architecture; 3) TMRNet \cite{RN649}, a memory bank-based long-range temporal aggregation network; 4) TeCNO \cite{RN654}, a multi-stage TCN-based hierarchical refinement network; 5) OperA  \cite{RN651}, a transformer-based spatial feature aggregation network; 6) Trans-SVNet \cite{jin2022trans}, a transformer-based spatiotemporal aggregation network. For ResNet-50, SV-RCNet, TeCNO, TMRNet, and Trans-SVNet, we implemented them using their open-sourced code. As for OperA, we implemented it based on the network architecture and settings described in the original paper. For a fair comparison, we used the limbus region-focused spatial feature for all two-stage-based comparison methods, namely TeCNO, OperA, and Trans-SVNet. Note that no additional supervision information, such as tool presence or phase remaining time, was employed in any of the methods.

Following \cite{RN660}, \cite{jin2022trans}, we evaluate the performance of surgical phase recognition using four metrics: accuracy (Acc), precision (Pre), recall (Rec) and Jaccard (Jac). The comparison results on both the Cataract-101 dataset and XH-CaTa dataset are listed in Table \ref{table2}. The results demonstrate that our LS-SAT network outperforms all the comparison methods. Notably, when compared to Trans-SVNet, the state-of-the-art method, our method achieves improvements with a margin of 1.7 percentage points (pp) for Acc, 2.6 pp for Pre, 2.7 pp for Rec, and 4.0 pp for Jac on the Cataract-101 dataset. Besides, our method achieves improvements with a margin of 2.3 pp for Acc, 2.0 pp for Pre, 3.9 pp for Rec, and 5.0 pp for Jac on the XH-CaTa dataset.

\begin{figure}[!t]
	\centerline{\includegraphics[width=\columnwidth]{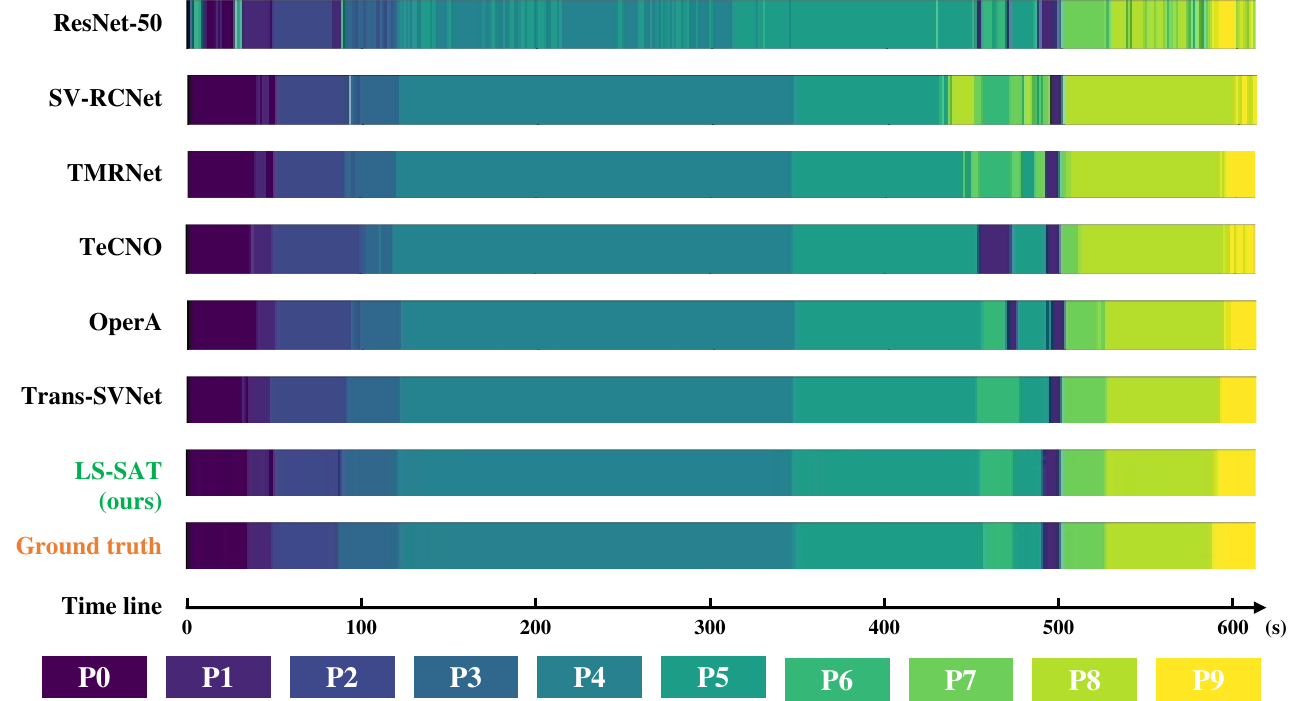}}
	\caption{Qualitative comparison results for a representative case in Cataract-101 dataset, using different surgical phase recognition methods. The surgical phase recognition results are visualized with color-coded ribbons.}
	\label{fig6}
\end{figure}

We provide qualitative comparison results using color-coded ribbons in Fig. \ref{fig6}, where each surgical phase is represented by a distinct color. The visualization reveals that our method can achieve a smoother surgical phase prediction and reduce the occurrence of instantaneous erroneous prediction frames.

\subsubsection{The effect of the limbus region-focused spatial feature extractor}
We performed an ablation study on the Cataract-101 dataset to evaluate the effectiveness of our proposed limbus region-focused spatial feature extractor. Specifically, we compared the segmentation and surgical phase recognition results obtained by our multi-task feature extractor network with those achieved by the single segmentation network and single surgical phase recognition network. The Dice score was used as the metric for limbus segmentation. Furthermore, we assessed the impact of the quality of the extracted spatial features on the spatiotemporal aggregation stage.

The results are presented in Table \ref{table3}. We observe that the multi-task feature extractor network outperforms the single task-based methods in both surgical phase recognition and limbus segmentation. Moreover, utilizing the limbus region-focused spatial features led to improvements in the spatiotemporal aggregation stage, with an increase of 3.5 pp in Acc, 5.1 pp in Pre, 5.6 pp in Rec, and 7.9 pp in Jac index.

\begin{figure*}[!t]
	\centerline{\includegraphics[width=\textwidth]{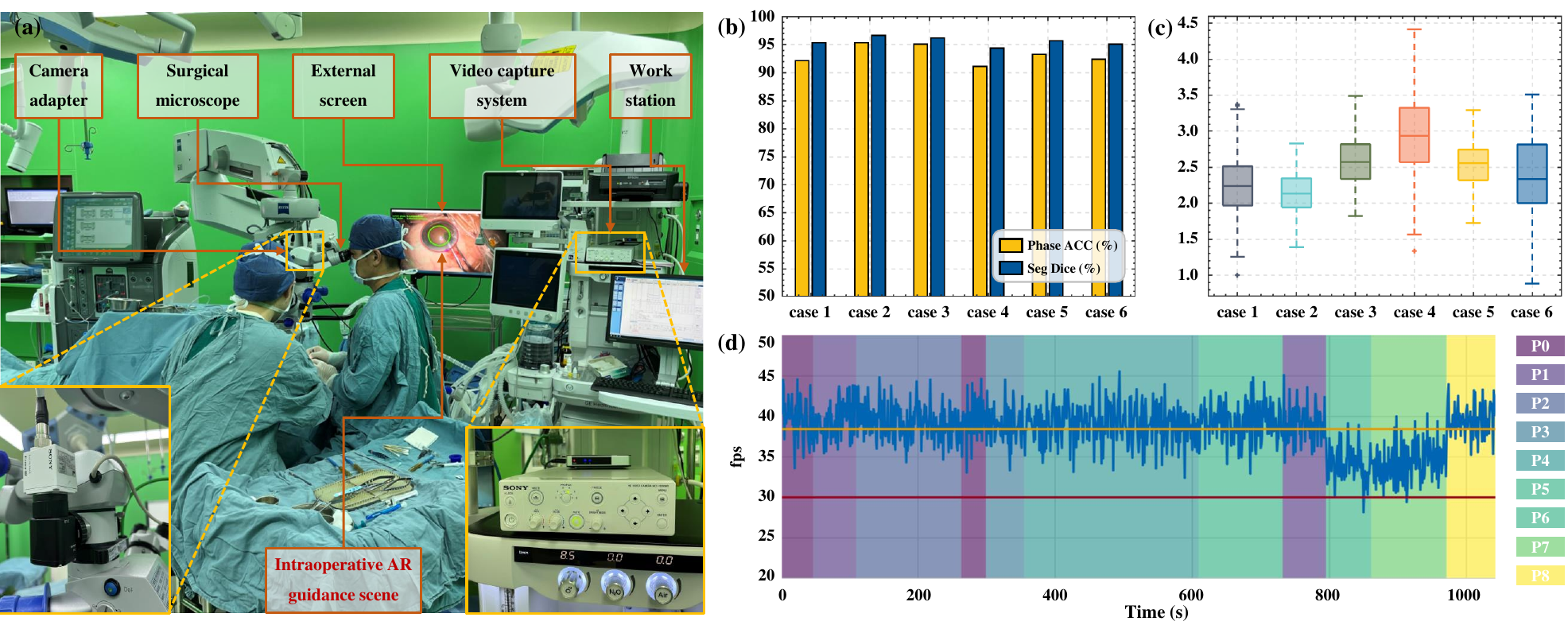}}
	\caption{Clinical experimental setup and performance evaluation results. (a) Overview of the clinical experimental setup and the hardware components integrated into our AR guidance system for PCS. (b) The results of phase recognition accuracy and segmentation Dice scores on six clinical cases. (c) The results of the rotation error evaluation across the same six clinical cases. Box plots are used to visualize the upper and lower bounds, along with the average values of the rotation error. Outliers are marked with a cross symbol. (d) Temporal changes in AR visualization fps for case 1. Each surgical phase is highlighted with a distinct background color. The average output fps is represented in orange, while the input fps from the camera is denoted in red.}
	\label{fig2}
\end{figure*}

\begin{figure*}[t]
	\centerline{\includegraphics[width=\textwidth]{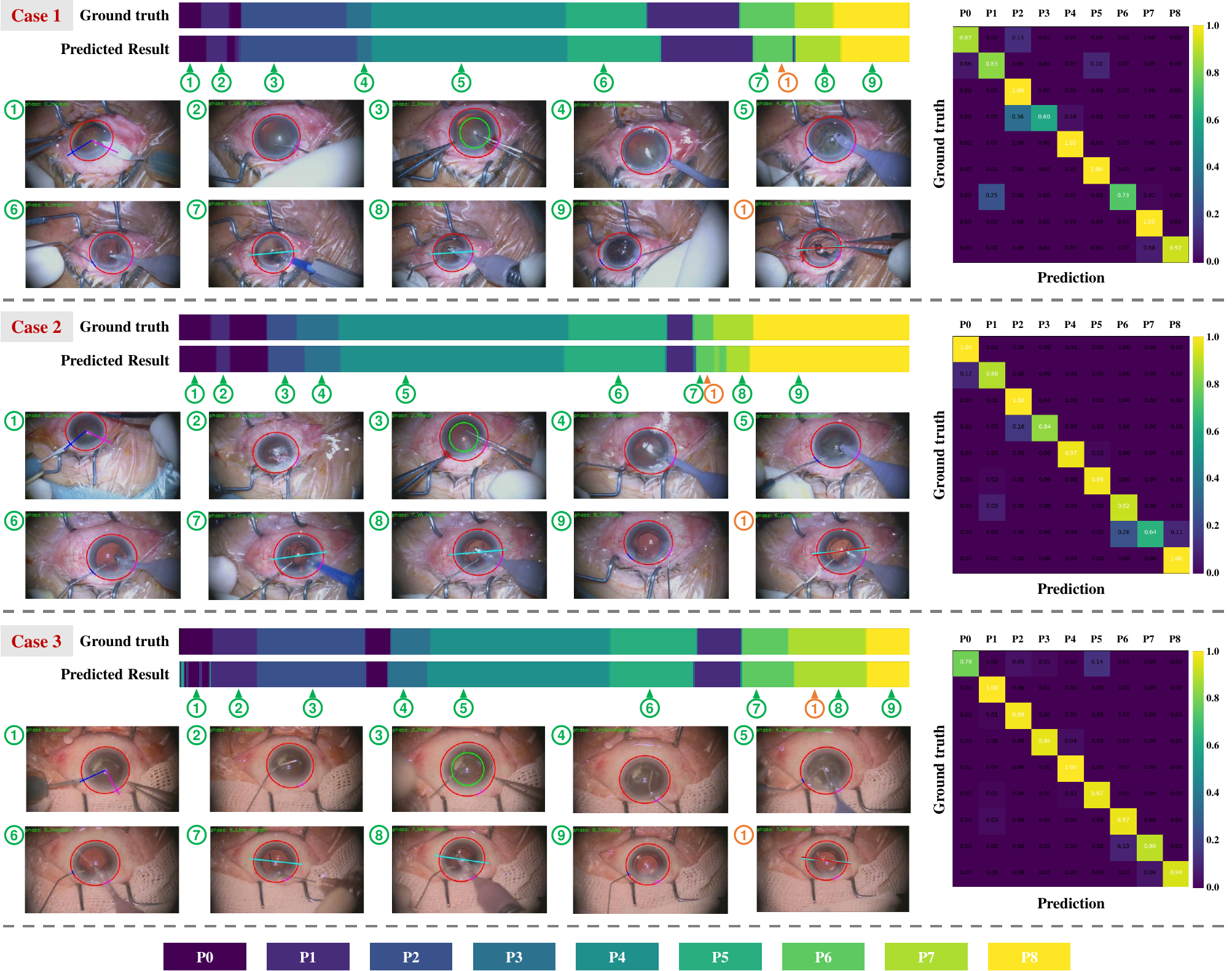}}
	\caption{Intraoperative AR scenes for three representative clinical cases. Each case displays visual representations of predicted surgical phase results and the corresponding ground truth using color-coded ribbons. These scenes highlight distinctive AR visuals for nine surgical phases, indicated by green markers. Additionally, a single frame from either P6 or P7 demonstrates the computed RRL represented by a light-blue line, as well as the ground truth RRL indicated by a red line. On the right, we present the results of online surgical phase recognition through a confusion matrix.}
	\label{fig11}
\end{figure*}

\subsubsection{The effect of the long-short spatiotemporal aggregation}
We conducted an ablation study on the Cataract-101 dataset to evaluate the effect of different combinations of long-range temporal aggregation, short-range temporal aggregation, and spatial feature aggregation. The quantitative results, presented in Table \ref{table4}, highlight that the method combining long features and short features outperforms the methods using each of them individually. Additionally, the incorporation of spatial features in both the long features-based method and the short features-based method yields improved results. Notably, our proposed structure, which combines long-short temporal features and spatial features, achieves the best performance when compared to other combinations. 

\subsection{Clinical evaluation of the AR guidance system}
\subsubsection{Experimental design and setup}
The clinical setup and hardware components of our developed intraoperative AR guidance system for PCS is shown in Fig. \ref{fig2} (a). A surgical microscope (OPMI Lumera T, Carl Zeiss Meditec AG, Germany) is employed for ophthalmic surgeries. To enable seamless video streaming, a camera adapter (3CMOS, Sony Corporation, Japan) is mounted on the microscope, connecting it to the surgical video capture system (MCC-1000MD, Sony Corporation, Japan). The captured intraoperative video is transmitted to a workstation (Precision 5820-Tower, Dell Technologies Inc., USA), where our proposed spatiotemporal network is deployed for real-time surgical phase recognition and limbus region segmentation. Subsequently, the extracted limbus region is used to compute the guidance parameters and the eye rotation parameters. This information contributes to design AR visual cues and construct a phase-specific AR scene, promptly displayed on an external screen (T3252U, HKC Corporation, China).

After approval by the ethics committee of the Xinhua Hospital Affiliated to Shanghai Jiao Tong University School of Medicine (Approval No. XHEC-D-2023-130), we conducted preliminary clinical experiments to evaluate the performance of our developed AR guidance system for PCS. The study involved six clinical cases conducted by a senior ophthalmologist. It is essential to highlight that this study adheres to the standard of sterile operation for ophthalmic surgery. Furthermore, adhering to clinical ethical boundaries, the implementation of the hardware equipment does not interfere with the traditional surgical field or the established practices of the ophthalmologists. In other words, the system was not actively guiding the ophthalmologist's actions but rather operated in sync and displayed information on an external screen for the sole purpose of algorithm evaluation.

To mitigate the domain-shift problem associated with deep learning algorithms, the spatiotemporal learning model utilized in the clinical experiment was trained on the XH-CaTa dataset. We evaluated the accuracy and real-time performance of the AR guidance system and analyzed instances of typical failed AR scenes. The results are described below.

\subsubsection{Accuracy evaluation}
We evaluate the accuracy of the developed AR guidance system for PCS from three aspects: online surgical phase recognition accuracy, limbus segmentation accuracy, and eye rotation accuracy.

For each clinical case, we recorded both the original video and AR guidance video. The surgical phase and limbus region in the original videos were manually annotated by an ophthalmologist with fps of 1, serving as the ground truth for microscope video recognition. We compared the intraoperative prediction results in the AR guidance video with the ground truth and calculated the surgical phase recognition and segmentation accuracy. Additionally, following \cite{RN648}, we manually annotated evident conjunctival vascular bifurcation landmarks, which were used in conjunction with the limbus central point to compute the rotation ground truth. We compared the predicted rotation results with the ground truth and computed the mean rotation error. 

We present the surgical phase recognition and segmentation Dice results for six clinical cases in Fig. \ref{fig2} (b). Our developed system consistently demonstrates accurate and stable surgical phase recognition and limbus segmentation, with an average surgical phase recognition accuracy of 93.3 ± 1.8 and a mean limbus segmentation Dice score of 95.6 ± 0.7. Additionally, we display the mean rotation error in P7 and P8 for all six cases in Fig. \ref{fig2} (c). It's noteworthy that, with the exception of case 4, the upper bounds of all other cases are below 3.5 mm, highlighting the system's potential for guiding the placement of the IOL.

In Fig. \ref{fig11}, we present the qualitative results from three representative cases. These illustrations include color-coded ribbons that convey the overall performance in surgical phase recognition, along with specific AR scenes at different surgical phases, which are indicated by green markers. Additionally, we include frames displaying the computed RRL alongside the ground truth RRL for each case, denoted by orange markers. Moreover, we depict the confusion matrix across the nine surgical phases for each case. Our findings demonstrate that in the online deployment of our system within a clinical setup, it consistently delivers accurate and smooth surgical phase predictions, along with precise phase-specific AR guidance.

\begin{figure*}[t]
	\centerline{\includegraphics[width=\textwidth]{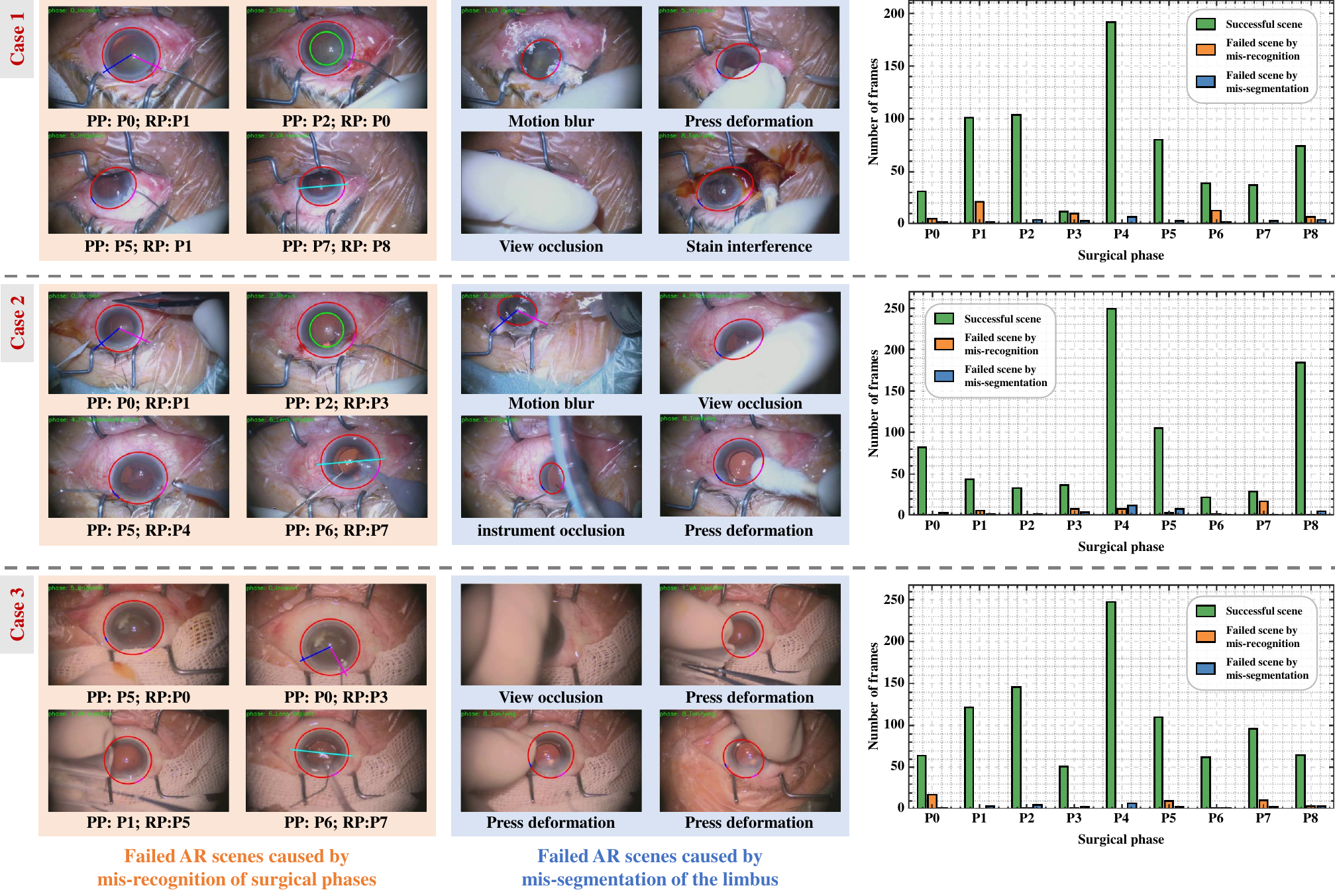}}
	\caption{Failed AR scenes in three selected clinical cases. In each case, AR scenes resulting from mis-recognition of surgical phases are highlighted with an orange background. We also specify their predicted phase (PP) and the real phase (RP). Additionally, AR scenes affected by mis-segmentation of the limbus are marked with a light-blue background, along with an explanation of the mis-segmentation cause. On the right side, we depict the number of successful and failed frames across various surgical phases.}
	\label{fig12}
\end{figure*}

\subsubsection{Real-time performance evaluation} 
We recorded the computing time consumption of each frame, including two-stage spatiotemporal aggregation network inference time, intraoperative AR guidance parameter computation time, and AR visualization time, and calculated the mean fps of each AR guidance video.

In Fig. \ref{fig2} (d), we present the temporal variation of fps across different surgical phases for case 1. The results reveal that the frame rates consistently remain above 30 fps for the majority of the time, which aligns with the online deployment requirements, as the video capture system maintains an output of 30 fps. However, there are instances, primarily during P6 (Lens implant) and P7 (VA removal), where the fps temporarily drops below 30. Quantitatively, the mean fps during P6 and P7 is 34.2 ± 2.2, while in other surgical phases, it averages 39.3 ± 2.3. These variations are expected, as we compute the eye rotation parameters in P6 and P7 but not in other surgical phases.

In clinical practice, we maintained a consistent visualization fps of 30 on the external screen to ensure stable and smooth AR guidance. Frames with a rate below 30 were omitted during visualization, without compromising the overall display stability.

\subsubsection{Failed AR scenes analysis}
We showcase typical instances of failed AR scenes in Fig. \ref{fig12}. These failures can be primarily attributed to two factors: mis-recognition of surgical phases or mis-segmentation of the limbus. In situations where mis-recognition of surgical phases occurs, these errors are particularly prevalent during phase transitions, leading to the display of visual cues that are inappropriate for the given phase. On the other hand, mis-segmentation of the limbus, a common issue during failed scenes, can be triggered by factors such as motion blur, pressure-induced deformation, view obstruction, instrument interference, and staining interference. This mis-segmentation leads to miscalculated intraoperative AR guidance parameters, ultimately resulting in the misalignment of visual cues.

We tallied the instances of both failed and successful AR scenes for every surgical phase in each case and have presented the results on the right side of Fig. \ref{fig12}. Our analysis reveals that, for each surgical phase, the successful scenes constitute the vast majority, demonstrating the effective implementation of our system in clinical practice. Additionally, we observed that AR scenes resulting from mis-segmentation issues are exceedingly rare.

\section{Discussion and conclusion}

In this study, we developed a novel phase-specific intraoperative AR guidance system for PCS. Our system incorporated a spatiotemporal learning network for surgical phase recognition and limbus region segmentation, which were subsequently utilized to calculate the parameters of intraoperative visual cues. By superimposing distinctive AR visual cues corresponding to the recognized surgical phase onto the microscopic video, our system has the potential to enhance ophthalmologists' intraoperative skills.

The experimental results on two datasets demonstrated that our proposed online surgical phase recognition method achieved smoother surgical phase prediction results compared to frame-wise methods \cite{RN665} and short-range temporal aggregation networks \cite{RN660}, \cite{RN649}. This outcome is highly advantageous for the implementation of AR guidance as it prevents potential interference to ophthalmologists resulting from incorrect superposition of visual cues caused by frequent misrecognition of surgical phases. Our method outperformed other long-range temporal aggregation networks \cite{RN654} \cite{RN651} \cite{jin2022trans} in handling challenging frames, particularly during phase conversion. This elevated effectiveness can be attributed to the combination of short-range temporal features, long-range temporal features, and spatial features. We implemented our spatiotemporal network in a clinical setting to assess the accuracy and real-time performance of our developed AR guidance system for PCS. Our findings indicated that both mis-recognition of surgical phases and mis-segmentation of the limbus can result in failed AR scenes. Specifically, these failed scenes commonly occurred when the limbus region was obstructed or when the video was blurred due to rapid eye or microscope motion. However, it is noteworthy that these erroneous scenarios have limited impact on ophthalmologists as they pertain to non-surgical instrument operation situations.

Nevertheless, our work has some limitations. Firstly, the diversity of surgical instruments and variations in surgical phases across different ophthalmic centers present a challenge in generalizing a model trained on a dataset from one clinical center to be applicable in other clinical centers. In this study, we specifically used the model trained on the XH-CaTa dataset for the clinical validation. By focusing on a single dataset, we aim to provide a more reliable evaluation of our approach in a consistent and controlled setting. Secondly, the use of supervised learning for anatomical segmentation requires manual annotation of the limbus region for each frame sampled from the dataset, which is a labor-intensive task.

In future work, we plan to enhance the AR visualization by directly integrating an AR display device \cite{RN656} into the eyepieces of the surgical microscope. Additionally, we plan to incorporate tracking and pose estimation of surgical instruments to enable intraoperative AR surgical navigation. Moreover, the application of our AR guidance system, whether through an external display or a direct overlay on the eyepiece, in guiding ophthalmologists during PCS can provide a comprehensive evaluation of its clinical feasibility and potential for widespread adoption.

\bibliographystyle{ieeetr}
\bibliography{refs}

\begin{thebibliography}{10}

\bibitem{RN661}
K.~Schoeffmann, M.~Taschwer, S.~Sarny, B.~Muenzer, M.~J. Primus, D.~Putzgruber,
  and M.~Assoc~Comp, ``Cataract-101-video dataset of 101 cataract surgeries,''
  in {\em 9th ACM Multimedia Systems Conference (MMSys)}, pp.~421--425, 2018.

\bibitem{RN646}
J.-S. Lee, C.-H. Hou, and K.-K. Lin, ``Surgical results of phacoemulsification
  performed by residents: A time-trend analysis in a teaching hospital from
  2005 to 2021,'' {\em Journal of Ophthalmology}, vol.~2022, 2022.

\bibitem{RN645}
Z.-L. Ni, G.-B. Bian, Z.~Li, X.-H. Zhou, R.-Q. Li, and Z.-G. Hou, ``Space
  squeeze reasoning and low-rank bilinear feature fusion for surgical image
  segmentation,'' {\em IEEE Journal of Biomedical and Health Informatics},
  vol.~26, no.~7, pp.~3209--3217, 2022.

\bibitem{RN658}
H.~Al~Hajj, M.~Lamard, P.-H. Conze, S.~Roychowdhury, X.~Hu, G.~Marsalkaite,
  O.~Zisimopoulos, M.~A. Dedmari, F.~Zhao, J.~Prellberg, M.~Sahu, A.~Galdran,
  T.~Araujo, V.~Duc~My, C.~Panda, N.~Dahiya, S.~Kondo, Z.~Bian, A.~Vandat,
  J.~Bialopetravicius, E.~Flouty, C.~Qiu, S.~Dill, A.~Mukhopadhyay, P.~Costa,
  G.~Aresta, S.~Ramamurthys, S.-W. Lee, A.~Campilho, S.~Zachow, S.~Xia,
  S.~Conjeti, D.~Stoyanov, J.~Armaitis, P.-A. Heng, W.~G. Macready,
  B.~Cochener, and G.~Quellec, ``Cataracts: Challenge on automatic tool
  annotation for cataract surgery,'' {\em Medical Image Analysis}, vol.~52,
  pp.~24--41, 2019.

\bibitem{RN648}
Y.~Zhai, G.~Zhang, L.~Zheng, G.~Yang, K.~Zhao, Y.~Gong, Z.~Zhang, X.~Zhang,
  B.~Sun, and Z.~Wang, ``Computer-aided intraoperative toric intraocular lens
  positioning and alignment during cataract surgery,'' {\em Ieee Journal of
  Biomedical and Health Informatics}, vol.~25, no.~10, pp.~3921--3932, 2021.

\bibitem{RN653}
L.~Ma and B.~Fei, ``Comprehensive review of surgical microscopes: technology
  development and medical applications,'' {\em Journal of Biomedical Optics},
  vol.~26, no.~1, 2021.

\bibitem{RN657}
F.~Yu, G.~S. Croso, T.~S. Kim, Z.~Song, F.~Parker, G.~D. Hager, A.~Reiter,
  S.~S. Vedula, H.~Ali, and S.~Sikder, ``Assessment of automated identification
  of phases in videos of cataract surgery using machine learning and deep
  learning techniques,'' {\em Jama Network Open}, vol.~2, no.~4, 2019.

\bibitem{RN647}
R.~G. Nespolo, D.~Yi, E.~Cole, N.~Valikodath, C.~Luciano, and Y.~I. Leiderman,
  ``Evaluation of artificial intelligence-based intraoperative guidance tools
  for phacoemulsification cataract surgery,'' {\em Jama Ophthalmology},
  vol.~140, no.~2, pp.~170--177, 2022.

\bibitem{RN660}
Y.~Jin, Q.~Dou, H.~Chen, L.~Yu, J.~Qin, C.-W. Fu, and P.-A. Heng, ``Sv-rcnet:
  Workflow recognition from surgical videos using recurrent convolutional
  network,'' {\em Ieee Transactions on Medical Imaging}, vol.~37, no.~5,
  pp.~1114--1126, 2018.

\bibitem{RN654}
T.~Czempiel, M.~Paschali, M.~Keicher, W.~Simson, H.~Feussner, K.~Seong~Tae, and
  N.~Navab, {\em TeCNO: surgical phase recognition with multi-stage temporal
  convolutional networks}.
\newblock Medical Image Computing and Computer Assisted Intervention - MICCAI
  2020. 23rd International Conference. Proceedings. Lecture Notes in Computer
  Science, 2020.

\bibitem{RN651}
T.~Czempiel, M.~Paschali, D.~Ostler, S.~T. Kim, B.~Busam, and N.~Navab,
  ``Opera: Attention-regularized transformers for surgical phase recognition,''
  in {\em International Conference on Medical Image Computing and Computer
  Assisted Intervention (MICCAI)}, vol.~12904 of {\em Lecture Notes in Computer
  Science}, pp.~604--614, 2021.

\bibitem{jin2022trans}
Y.~Jin, Y.~Long, X.~Gao, D.~Stoyanov, Q.~Dou, and P.-A. Heng, ``Trans-svnet:
  hybrid embedding aggregation transformer for surgical workflow analysis,''
  {\em International Journal of Computer Assisted Radiology and Surgery},
  vol.~17, no.~12, pp.~2193--2202, 2022.

\bibitem{RN642}
W.~Yue, H.~Liao, Y.~Xia, V.~Lam, J.~Luo, and Z.~Wang, ``Cascade multi-level
  transformer network for surgical workflow analysis,'' {\em IEEE transactions
  on medical imaging}, vol.~42, no.~10, pp.~2817--2831, 2023.

\bibitem{RN659}
F.~Yi and T.~Jiang, ``Hard frame detection and online mapping for surgical
  phase recognition,'' in {\em 10th International Workshop on Machine Learning
  in Medical Imaging (MLMI) / 22nd International Conference on Medical Image
  Computing and Computer-Assisted Intervention (MICCAI)}, vol.~11768 of {\em
  Lecture Notes in Computer Science}, pp.~449--457, 2019.

\bibitem{RN655}
Y.~Jin, H.~Li, Q.~Dou, H.~Chen, J.~Qin, C.-W. Fu, and P.-A. Heng, ``Multi-task
  recurrent convolutional network with correlation loss for surgical video
  analysis,'' {\em Medical Image Analysis}, vol.~59, 2020.

\bibitem{yi2022not}
F.~Yi, Y.~Yang, and T.~Jiang, ``Not end-to-end: Explore multi-stage
  architecture for online surgical phase recognition,'' in {\em Proceedings of
  the Asian Conference on Computer Vision}, pp.~2613--2628, 2022.

\bibitem{RN662}
S.~Drouin, A.~Kochanowska, M.~Kersten-Oertel, I.~J. Gerard, R.~Zelmann,
  D.~De~Nigris, S.~Beriault, T.~Arbel, D.~Sirhan, A.~F. Sadikot, J.~A. Hall,
  D.~S. Sinclair, K.~Petrecca, R.~F. DelMaestro, and D.~L. Collins, ``Ibis: an
  or ready open-source platform for image-guided neurosurgery,'' {\em
  International Journal of Computer Assisted Radiology and Surgery}, vol.~12,
  no.~3, pp.~363--378, 2017.

\bibitem{RN666}
I.~Cabrilo, K.~Schaller, and P.~Bijlenga, ``Augmented reality-assisted bypass
  surgery: Embracing minimal invasiveness,'' {\em World Neurosurgery}, vol.~83,
  no.~4, pp.~596--602, 2015.

\bibitem{roodaki2015introducing}
H.~Roodaki, K.~Filippatos, A.~Eslami, and N.~Navab, ``Introducing augmented
  reality to optical coherence tomography in ophthalmic microsurgery,'' in {\em
  2015 IEEE international symposium on mixed and augmented reality}, pp.~1--6,
  IEEE, 2015.

\bibitem{pan2020real}
J.~Pan, W.~Liu, P.~Ge, F.~Li, W.~Shi, L.~Jia, and H.~Qin, ``Real-time
  segmentation and tracking of excised corneal contour by deep neural networks
  for dalk surgical navigation,'' {\em Computer Methods and Programs in
  Biomedicine}, vol.~197, p.~105679, 2020.

\bibitem{bian2023variation}
G.-B. Bian, W.-Q. Yue, Z.~Li, L.~Zhang, S.~Zhang, W.-P. Liu, S.~Li, E.~P.
  Medeiros, W.-Q. Wu, and V.~H.~C. de~Albuquerque, ``Variation-learning
  high-resolution network for capsulorhexis recognition of cataract surgery,''
  {\em Applied Soft Computing}, p.~110841, 2023.

\bibitem{RN641}
R.~G. Nespolo, D.~Yi, E.~Cole, D.~Wang, A.~Warren, and Y.~I. Leiderman,
  ``Feature tracking and segmentation in real time via deep learning in
  vitreoretinal surgery a platform for artificial intelligence-mediated
  surgical guidance,'' {\em Ophthalmology Retina}, vol.~7, no.~3, pp.~236--242,
  2023.

\bibitem{RN663}
A.~P. Twinanda, S.~Shehata, D.~Mutter, J.~Marescaux, M.~de~Mathelin, and
  N.~Padoy, ``Endonet: A deep architecture for recognition tasks on
  laparoscopic videos,'' {\em Ieee Transactions on Medical Imaging}, vol.~36,
  no.~1, pp.~86--97, 2017.

\bibitem{wang2022intelligent}
T.~Wang, J.~Xia, R.~Li, R.~Wang, N.~Stanojcic, J.-P.~O. Li, E.~Long, J.~Wang,
  X.~Zhang, J.~Li, {\em et~al.}, ``Intelligent cataract surgery supervision and
  evaluation via deep learning,'' {\em International Journal of Surgery},
  vol.~104, p.~106740, 2022.

\bibitem{RN644}
X.~Zou, W.~Liu, J.~Wang, R.~Tao, and G.~Zheng, ``Arst: auto-regressive surgical
  transformer for phase recognition from laparoscopic videos,'' {\em Computer
  Methods in Biomechanics and Biomedical Engineering-Imaging and
  Visualization}, vol.~11, no.~4, pp.~1012--1018, 2023.

\bibitem{yue2023cascade}
W.~Yue, H.~Liao, Y.~Xia, V.~Lam, J.~Luo, and Z.~Wang, ``Cascade multi-level
  transformer network for surgical workflow analysis,'' {\em IEEE Transactions
  on Medical Imaging}, 2023.

\bibitem{RN665}
K.~He, X.~Zhang, S.~Ren, J.~Sun, and Ieee, ``Deep residual learning for image
  recognition,'' in {\em 2016 IEEE Conference on Computer Vision and Pattern
  Recognition (CVPR)}, IEEE Conference on Computer Vision and Pattern
  Recognition, pp.~770--778, 2016.

\bibitem{RN668}
O.~Ronneberger, P.~Fischer, and T.~Brox, ``U-net: Convolutional networks for
  biomedical image segmentation,'' in {\em 18th International Conference on
  Medical Image Computing and Computer-Assisted Intervention (MICCAI)},
  vol.~9351 of {\em Lecture Notes in Computer Science}, pp.~234--241, 2015.

\bibitem{RN664}
A.~Vaswani, N.~Shazeer, N.~Parmar, J.~Uszkoreit, L.~Jones, A.~N. Gomez,
  L.~Kaiser, and I.~Polosukhin, ``Attention is all you need,'' in {\em 31st
  Annual Conference on Neural Information Processing Systems (NIPS)}, vol.~30
  of {\em Advances in Neural Information Processing Systems}, 2017.

\bibitem{xu2021long}
M.~Xu, Y.~Xiong, H.~Chen, X.~Li, W.~Xia, Z.~Tu, and S.~Soatto, ``Long
  short-term transformer for online action detection,'' {\em Advances in Neural
  Information Processing Systems}, vol.~34, pp.~1086--1099, 2021.

\bibitem{RN667}
D.~Eigen, R.~Fergus, and Ieee, ``Predicting depth, surface normals and semantic
  labels with a common multi-scale convolutional architecture,'' in {\em IEEE
  International Conference on Computer Vision}, IEEE International Conference
  on Computer Vision, pp.~2650--2658, 2015.

\bibitem{RN643}
W.~Zhao, Z.~Zhang, Z.~Wang, Y.~Guo, J.~Xie, and X.~Xu, ``Eclnet: Center
  localization of eye structures based on adaptive gaussian ellipse heatmap,''
  {\em Computers in Biology and Medicine}, vol.~153, 2023.

\bibitem{RN650}
T.~Xia and F.~Jia, ``Against spatial-temporal discrepancy: contrastive
  learning-based network for surgical workflow recognition,'' {\em
  International Journal of Computer Assisted Radiology and Surgery}, vol.~16,
  no.~5, pp.~839--848, 2021.

\bibitem{RN669}
J.~Deng, W.~Dong, R.~Socher, L.-J. Li, K.~Li, F.-F. Li, and Ieee, ``Imagenet: A
  large-scale hierarchical image database,'' in {\em IEEE-Computer-Society
  Conference on Computer Vision and Pattern Recognition Workshops}, IEEE
  Conference on Computer Vision and Pattern Recognition, pp.~248--255, 2009.

\bibitem{RN649}
Y.~Jin, Y.~Long, C.~Chen, Z.~Zhao, Q.~Dou, and P.-A. Heng, ``Temporal memory
  relation network for workflow recognition from surgical video,'' {\em Ieee
  Transactions on Medical Imaging}, vol.~40, no.~7, pp.~1911--1923, 2021.

\bibitem{RN656}
P.-H.~C. Chen, K.~Gadepalli, R.~MacDonald, Y.~Liu, S.~Kadowaki, K.~Nagpal,
  T.~Kohlberger, J.~Dean, G.~S. Corrado, J.~D. Hipp, C.~H. Mermel, and M.~C.
  Stumpe, ``An augmented reality microscope with real-time artificial
  intelligence integration for cancer diagnosis,'' {\em Nature Medicine},
  vol.~25, no.~9, pp.~1453--+, 2019.

\end{thebibliography}

\end{document}